\pdfoutput=1

\documentclass[11pt]{article}

\usepackage[preprint]{acl}

\usepackage{times}
\usepackage{latexsym}

\usepackage[T1]{fontenc}

\usepackage[utf8]{inputenc}

\usepackage{microtype}
\usepackage{times}
\usepackage{url}
\usepackage{latexsym}

\usepackage{graphicx}
\graphicspath{ {./} }
\usepackage[table,xcdraw]{}
\usepackage{microtype}
\usepackage{booktabs}
\usepackage{amsmath}
\usepackage{color}
\usepackage{amssymb}
\usepackage{multirow} 
\usepackage{makecell}
\usepackage{tabularx}
\usepackage{threeparttable}
\usepackage{amssymb}
\usepackage{marvosym}
\usepackage{subfigure}
\usepackage{soul}
\usepackage{times}
\usepackage{latexsym}
\usepackage{inconsolata}

\newcommand{\cursor}[1]{{\fontfamily{qcr}\selectfont #1}}

\title{Leveraging Hierarchical Prototypes as the Verbalizer for Implicit Discourse Relation Recognition} 

\author{Wanqiu Long\textsuperscript{1} \and Bonnie Webber\textsuperscript{1} \\
  \textsuperscript{1}University of Edinburgh, Edinburgh, UK  \\
  \texttt{Wanqiu.long@ed.ac.uk, Webber.Bonnie@ed.ac.uk}
}

\begin{document}

\maketitle

\begin{abstract}
Implicit discourse relation recognition involves determining relationships that hold between spans of text that are not linked by an explicit discourse connective. In recent years, the pre-train, prompt, and predict paradigm has emerged as a promising approach for tackling this task. However, previous work solely relied on manual verbalizers for implicit discourse relation recognition, which suffer from issues of ambiguity and even incorrectness. To overcome these limitations, we leverage the prototypes that capture certain class-level semantic features and the hierarchical label structure for different classes as the verbalizer. We show that our method improves on competitive baselines. Besides, our proposed approach can be extended to enable zero-shot cross-lingual learning, facilitating the recognition of discourse relations in languages with scarce resources. These advancement validate the practicality and versatility of our approach in addressing the issues of implicit discourse relation recognition across different languages. 
\end{abstract}

\section{Introduction}
Discourse relations play a crucial role in establishing coherence between sentences and clauses in a text. Automatically identifying the sense or senses of discourse relations between sentences and clauses is valuable for various downstream NLP tasks that rely on coherent understanding of text such as text summarization \cite{huang-kurohashi-2021-extractive}, question answer \cite{pyatkin-etal-2020-qadiscourse} and event relation extraction \cite{tang-etal-2021-discourse}.

Recently, for implicit discourse relation recognition, several work \cite{zhou-etal-2022-prompt-based,xiang-etal-2022-connprompt,Chan2023DiscoPromptPP} have applied prompt learning by transforming the task into a connective-cloze task. Figure 1 shows the main template and the verbalizer adopted in their work. The models need to infer the connectives in \cursor{[MASK]} and the filled connectives are mapped to corresponding labels via the mannually-designed verbalizer.

\begin{figure}[!tbp]
\setlength{\belowcaptionskip}{-8pt}

\centering

\includegraphics[height=2cm,width=7.8cm]{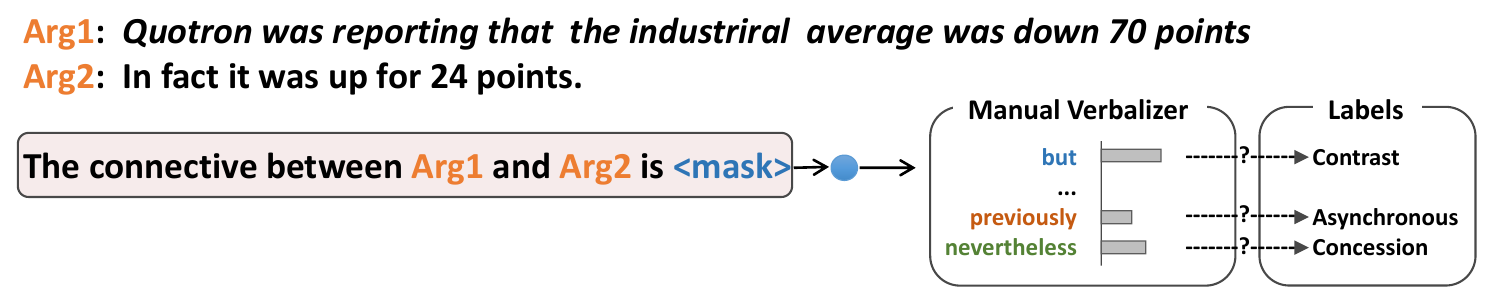}

\caption{The manual verbalizer and the labels.}
\label{fig1}
\vspace{-0.2cm}
\end{figure}

However, since most connectives are not only used for one sense, many connectives they selected can be connected with several labels instead of only one labels. In the given example depicted in Figure \ref{fig1}, the gold label of this example is ``Concession''. In their manual verbalizer, the connective ``but'' was connected with the label ``Contrast'' and ``nevertheless'' was mapped for the label ``Concession''. However, it should be noted that ``but'' can also be used to indicate other senses such as ``Concession'' and ``Substitution''(see Appendix C in PDTB-3 annotation manual \cite{webber2019penn}.) Likewise, ``nevertheless'' is not strictly limited to conveying ``Concession'' but can also be employed to indicate ``Contrast''. Similarly, the connective ``since'' is not exclusively associated with the label ``Pragmatic Cause'' but can also be used for ``Cause'' or ``Asynchronous''; The connective ``specifically'' can be used for ``List'' and``Restatement'', etc. Therefore, while previous approaches have shown improvements in task performance, the manual selection of connectives as verbalizers may not consistently serve as reliable indicators, resulting in potentially less accurate results for certain labels.

In this paper, instead of relying on manual design, our paper introduces an alternative verbalizer to implicit discourse relation recognition (IDRR) by leveraging the prototype learning and the sense hierarchy reflecting the organizations of the labels. We try to estimate the prototype vectors (the central points) for each class, which can then be utilized as verbalizer. We leverage contrastive learning \cite{pmlr-v119-chen20j} to adjust the distance among class prototypes, the distance among instances, and the distance between prototypes and instances based on the sense hierarchy. In the PDTB-2 \cite{prasad-etal-2008-penn} and PDTB-3 \cite{webber2019penn}, senses that can hold between adjacent spans are arranged in a three-level hierarchy, and the sense hierarchy has demonstrated its usefulness in the work \cite{wu2021label,long-webber-2022-facilitating,Chan2023DiscoPromptPP}. 

Moreover, the proposed approach not only works for monolingual IDRR but also facilitates zero-shot cross-lingual learning IDRR. Despite the impressive performance recently achieved by deep learning models for IDRR in recent years, the existing work primarily focus on a few languages with relatively rich annotated data, such as English. In contrast, most languages contain limited or no labeled data, constraining the application of existing methods to these languages. \citet{kurfali-ostling-2019-zero} present the first study on a zero-shot transfer learning for implicit discourse relation recognition by using English data and Turkish data, but there is no other work regarding transfer learning for the languages with few labeled data. Our approach can be extended to those data by using a cross-lingual template and a prototype alignment between the source language and the target languages.

The main contributions of our work are as follows:
\begin{itemize}
\setlength{\itemsep}{0pt}
\setlength{\parsep}{0pt}
\setlength{\parskip}{0pt}

    \item We leverage the hierarchical prototypes as the verbalizer for IDRR that eliminates the need for manual design and incorporates the sense hierarchy which can reflect the organization of discourse relational senses. 
    \item Our proposed approach can be effectively extended to enable zero-shot cross-lingual learning, facilitating the recognition of discourse relations in languages with limited resources. 
    \item The experimental results show that our approach shows improvement against competitive baselines for English and can effectively help other languages with limited data to get better results in the IDRR task. 

\end{itemize}

\section{Related Work}
\subsection{Verbalizer construction}
A Verbalizer is a core component of a prompt. The verbalizer is to connect the model outputs and labels \cite{zhao-etal-2023-pre}. Currently, there are mainly four types of verbalizers in terms of the construction: manual verbalizer\cite{schick-schutze-2021-exploiting}, soft verbalizer\cite{gao-etal-2021-making}, search-based verbalizer\cite{shin-etal-2020-autoprompt}, and prototypical verbalizer\cite{Cui2022PrototypicalVF}. \citet{Cui2022PrototypicalVF} proposed prototypical verbalizer for few-shot learning. It involves representing each class or label with a prototype vector to serve as a verbalizer. Following their work, our verbalizer is based on a hierarchy of prototypes to capture the essential information about the class and the hierarchical structure for implicit discourse relation recognition.

 
\subsection{Prototype-based learning}
\citet{snell2017prototypical} introduced prototypical networks. Prototype vectors are computed by taking the average of instance vectors, and predictions are made by comparing these prototypes to query instances using a metric-based approach. The prototype network has been applied to some downstream tasks such as emotion recognition \cite{song2022supervised}, relation extraction \cite{ding2021prototypical} and name entity recognition \cite{zhou2023improving}. Their success demonstrates prototypes, as representative embeddings of instances belonging to the same classes, capture certain class-level semantic features.

\subsection{Implicit Discourse Relation Recognition}

In the field of implicit discourse relation recognition, the pre-train and fine-tuning paradigm is commonly employed before \cite{shi-demberg-2019-next,Wu2021ALD,long-webber-2022-facilitating}. However, there has been a recent shift towards exploring prompt-based learning methods in this domain. Specifically, \citet{zhou-etal-2022-prompt-based} present a approach called prompt-based connective Prediction. \citet{xiang-etal-2022-connprompt} developed a connective-cloze prompt to transform the relation prediction task into a connective-cloze task. \citet{Chan2023DiscoPromptPP} incorporated the label dependencies information and connectives into pre-trained language models. As explicit connectives are linguistic expressions that explicitly signal the relationship between two adjacent discourse segments, all these work utilize the correlations between the connectives and the sense labels, manually selecting connectives as the verbalizers to connect the outputs of model and the labels. However, most of the connectives are not exclusive to only one sense label. Therefore, the manual verbalizers might introduce potential inaccuracies. In our work, instead of relying on manual verbalizers, we propose the use of prototypes as the verbalizer to address this limitation.

Furthermore, it is important to note that the aforementioned studies have primarily focused on English datasets. Only \citet{kurfali-ostling-2019-zero} presented a system that utilized zero-shot transfer learning for implicit discourse relation classification on datasets in other languages. Our work goes beyond the confines of English and incorporates other languages with limited annotated data. We achieve this by extending our approach to the zero-shot cross-lingual transfer learning scenario.



\section{Methodology}
\label{sec3}
\subsection{Background}
\subsubsection{Supervised Contrastive Loss}
Supervised contrastive loss can leverage label information more effectively, as it can pull together the clusters of points belonging to the same class and push apart clusters of samples from different classes \cite{khosla2020supervised}. 

To calculate the contrastive loss, an input batch is $x_k$, and $z_{2k}$ and $z_{2k-1}$ are the features of the two augmented views of the input batch $x_k$. The supervised contrastive learning loss can be computed as
\begin{equation}
    \mathcal{L}_{scl}=Scl(z_{2k-1},z_{2k}, y_k) 
\end{equation}

In this equation, $y_k$ denotes the label for batch $x_k$. More details of the supervised learning loss can be found in \citet{khosla2020supervised}.


\subsubsection{Prompt-based Tuning}
The original prompt-based tuning approach transforms the downstream task into a cloze question with masks.. An example is shown in Figure 1. In order to create the prompt input, the input x is added a template T(·) = ``\cursor{The connective between Arg1 and Arg2 is [MASK]}''. The verbalizer maps labels to connectives that can indicate the labels. For examples,
The discourse relation ``Asynchronous'' is mapped as the connective ``previously''.



\begin{figure*}[!tbp]
\setlength{\belowcaptionskip}{-8pt}

\centering

\includegraphics[width = 13cm
]{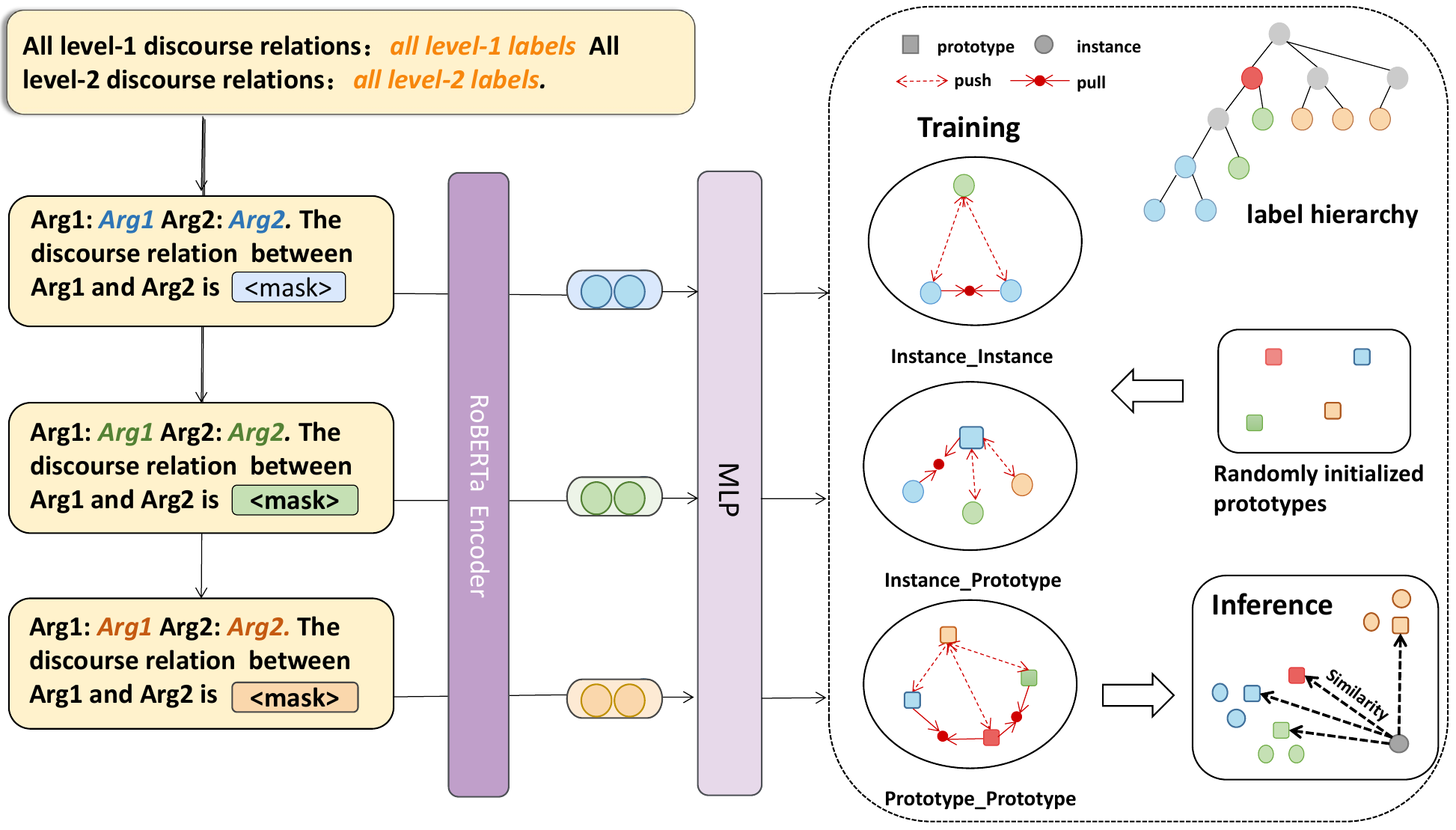}

\caption{The hidden states of \cursor{[MASK]} token represent instances and project them to another embedding space for prototype learning. Three contrastive learning losses adjust the distances among prototypes, the distances among instances, and the distances between prototypes and instances based on the sense hierarchy. Finally, we calculate the similarity scores of query and prototypes during inference. }
\vspace{-0.1cm}
\end{figure*}

\subsection{Our Approach}
Figure 2 illustrates our framework. We employ a template where Arg1 and Arg2, representing two discourse segments, are concatenated into a single word sequence. To provide background information, we insert the label information at both the top level and the second level at the beginning of Arg1 and Arg2. The template is to predict implicit discourse relations between Arg1 and Arg2.

Then,  following \citet{Cui2022PrototypicalVF}, instead of designing manual verbalizer or using the label words as the answer set for the \cursor{[MASK]} token, we use the hidden states of \cursor{[MASK]} token to represent instances, then project them to another embedding space for prototype learning. The prototypes serve as the verbalizer for prediction. Given a piece of training text $x$ wrapped with a template, we take the last layer’s hidden state of the \cursor{[MASK]} token $\text{ h}_{\texttt{\cursor{[MASK]}}}$ as the initial representation of the text. With an encoder $E_\phi(x)$ parameterized by $\phi$, the instance representation of $x$ is:
\begin{equation}
    v = E_\phi(x) =W\text{h}_{\texttt{\cursor{[mask]}}}
\end{equation}

We randomly initialize a vector to represent a category. Denote $C = \{c_1, ... , c_M \}$ as the set of prototype vectors. We just adopt a linear layer with weight \textbf{W}.


We leverage contrastive learning to adjust the distances among class prototypes, the distances among instances, and the distances between prototypes and instances in terms of the sense hierarchy. 

(1) For instance to instance pairs, intra-class pairs should get higher similarity scores than inter-class pairs:
 \begin{equation}
  \begin{split}
    \mathcal{L}_{ins\_ins}=-\frac{1}{N}\sum_{i=1}^N\frac{1}{|e^i_{pos}|}\sum_{j=1}^{N}1_{i\ne j}1_{j\in e^i_{pos}}\\ log\frac{e^{sim(v_j,v_i)/\tau}}{\sum_{k=1}^{N}1_{i\ne k}e^{sim(v_k,v_i)/\tau}}
  \end{split}
 \end{equation}
where $e_{pos}^i$ represents the set of positive examples for the $i$-th instance in the same batch. For every instance in a batch, any one of other examples in this batch can form pairs with. Therefore, if the batch size is $N$, we will have $N^2-1$ pairs in total.
 
We follow \citet{long-webber-2022-facilitating} to narrow the distances among examples from the same types at level-2 or level-3 and enlarge the distances among examples from different types at level-2 or level-3 in terms of the sense hierarchy.

(2) For instance to prototype pairs, the loss function is to force each prototype to lie at the center point of its instances:

 \begin{equation}
  \begin{split}
    \mathcal{L}_{ins\_pro}=-\frac{1}{N}\sum_{i=1}^Nlog\frac{e^{sim(v_i,c_i)/\tau}}{\sum_{j=1}^{M}e^{sim(v_i,c_j)/\tau}}
  \end{split}
 \end{equation}
where $c_i$ represents the prototype of the corresponding category of the example $v_i$ . We create pairs by pairing each prototype of a class with examples from the same batch. For a specific class, the number of pairs will be $N$. For the instance-prototype pairs, each example has three levels, so each example has three prototypes. This is to narrow the distances between the example and the prototype of its class and to widen the distances between the example and prototypes of other classes.

(3) For the prototype to prototype pairs, this loss function maximizes intra-class similarity and minimizes inter-class similarity between prototypes. In terms of the sense hierarchy, a father prototype should be nearer from its children prototype. We pair each father prototype with each child prototype, resulting in a total number of pairs equal to $N_1 \times N_2$, where $N_1$ is the number of father prototypes and $N_2$ is the number of child prototypes.

 \begin{equation}
  \begin{split}
    \mathcal{L}_{pro\_pro}=-\frac{1}{M}\sum_{i=1}^Mlog\frac{e^{sim(c_i,c_i')/\tau}}{\sum_{j=1}^{M'}e^{sim(c_i,c_j)/\tau}}
  \end{split}
 \end{equation}
where $c_i'$ represents the father class of $c_i$.

Overall, combining the instance-instance loss, instance-prototype loss and the prototype-prototype loss, the final training objective is:
\begin{equation}
    \mathcal{L}=\mathcal{L}_{ins\_ins}+\mathcal{L}_{ins\_pro}+\mathcal{L}_{pro\_pro}
\end{equation} 


\subsection{Inference}
During inference, we calculate the similarity scores of query and prototypes. For instance $i$, the probability score for class $k$ is 
\begin{equation}
    P(y_k|x) = \frac{e^{sim(v_i,c_i)}}{\sum_{j=1}^{M}e^{sim(v_i,c_j)}}. 
\end{equation}

To make predictions, we use the arg max function, selecting the label $y_k$ that maximizes the probability $P(y_k|x)$. We calculate the similarities between the representation of the examples and the level-1 prototypes for the level-1 senses prediction, while calculating the similarities between the representation of the examples and the level-2 prototypes for the level-2 senses prediction.

\section{Experiment setting}
In this section, we present our experiment settings, including the datasets, competitors, PLMs, and parameter settings.
\subsection{Datasets}
The recognition of discourse relations is often evaluated using the Penn Discourse TreeBank (PDTB) dataset. We used both PDTB-2 
 \cite{prasad-etal-2008-penn} and PDTB-3 \cite{webber2019penn} for evaluations.
Following earlier work \cite{bai-zhao-2018-deep,liu2020importancewordsentencerepresentation,xiang-etal-2022-encoding}, we adopt Sections 2-20 of the corpus for Training, Sections 0-1 for Validation, and Sections 21-22 for testing. In addition, following previous work\cite{shallow}, we treat the instances with multiple annotated labels as separate examples during training and a prediction matching one of the gold types is regarded as the correct answer at test time.

\subsection{Baselines}
We compare our model with the following state-of-the-art baselines to verify the effect of our approach.
Since most previous research evaluated on either the PDTB-2 or the PDTB-3 dataset, we compare different baselines specific to each dataset. 
\begin{table}
\centering
\tiny
\begin{tabular}{l|cc|cc}
\bottomrule
\multirow{3}{*}{Model}&\multicolumn{4}{c}{PDTB-2} \\

&\multicolumn{2}{c|}{Top Level}&\multicolumn{2}{c}{Second Level} \\

&Acc&Macro-F1&Acc&Macro-F1\\
\hline
\citet{dai-huang-2019-regularization}&59.66&52.89&48.23&33.41\\
\citet{shi-demberg-2019-learning}&61.42&46.40&47.83&-\\
\citet{Nguyen2019EmployingTC}&-&53.00&49.95&-\\
\citet{Guo2020WorkingMN}&57.25&47.90&-&-\\

\citet{Kishimoto2020AdaptingBT}&65.26&58.48&52.34&-\\
\citet{liu2020importancewordsentencerepresentation}&69.06&63.39&58.13&-\\
\citet{jiang-etal-2021-just}&-&57.18&-&37.76\\
\citet{dou-etal-2021-cvae-based}&70.17&65.06&-&-\\
\citet{Wu2022ALD}&71.18&63.73&60.33&40.49\\
\citet{long-webber-2022-facilitating}&71.70&67.85&59.19&45.54\\
\citet{zhou-etal-2022-prompt-based}&70.84&64.95&60.54&41.55 \\
\citet{Chan2023DiscoPromptPP}(T5-base)&71.70&65.79&\textbf{61.02}&43.68\\

\hline
Ours (RoBERTa-base)&\textbf{72.47}&\textbf{69.66}&60.73
&\textbf{47.07}
\\
\hline
\toprule
\end{tabular}
\caption
{Experimental results on PDTB-2.}
\end{table}
The baselines for PDTB-2 are: (1) a model which incorporates external event knowledge and coreference relations \cite{dai-huang-2019-regularization};(2) a neural model which learns better argument representations by utilizing the inserted connectives \cite{shi-demberg-2019-learning};(3)a model used to predict the labels and connectives at the same time \cite{Nguyen2019EmployingTC};(4)a working memory-driven neural networks by using a knowledge enhancement paradigm \cite{Guo2020WorkingMN};(5) a model which performs additional pre-training on text tailored to discourse classification \cite{Kishimoto2020AdaptingBT};(6)a RoBERTa-based model that combines a powerful contextualized representation module, a bilateral multi-perspective matching module, and a global information fusion module \cite{liu2020importancewordsentencerepresentation}; (7)a method that recognizes the relation label and generate the target sentence containing the meaning of relations simultaneously \cite{jiang-etal-2021-just}; (8) a method that develops a re-anchoring strategy by using Conditional VAE (CVAE) \cite{dou-etal-2021-cvae-based}; (9) a hierarchical contrastive learning based multi-task framework \cite{long-webber-2022-facilitating}; (10) a transformed prompt-based Connective prediction (PCP) task by utilizing the correlation between connectives \cite{zhou-etal-2022-prompt-based};(11) a method injecting the label dependencies information via prompt tuning with aligning the representations and using connectives prediction \cite{Chan2023DiscoPromptPP}. 

We compared the following baselines for PDTB-3: (1) a model with multi-level attention (NNMA) \cite{liu-li-2016-recognizing}; (2) a method which adopts a gated relevance network to capture the semantic interaction \cite{Chen2016DiscourseRD};
(3) a multi-task attention based neural network model through two types of representation learning  \cite{lan-etal-2017-multi}; (4)a propagative attention learning model using a cross-coupled two-channel network  \cite{ruan-etal-2020-interactively};
(5) a multi-Attentive Neural Fusion (MANF) model to encode and fuse both semantic connection and linguistic evidence for IDRR \cite{xiang-etal-2022-encoding};(6) a hierarchical contrastive learning based multi-task framework  \cite{long-webber-2022-facilitating};(7) a connective-cloze Prompt (ConnPrompt) to transform the relation prediction task as a connective-cloze task by designing insert-cloze Prompt and Prefix-cloze Prompt \cite{xiang-etal-2022-connprompt}.

\begin{table}
\setlength{\belowcaptionskip}{-0.1cm}
\centering
\scriptsize
\begin{tabular}{l|cc|cc}
\bottomrule
\multirow{3}{*}{Model}&\multicolumn{4}{c}{PDTB-3} \\

&\multicolumn{2}{c|}{Top Level}&\multicolumn{2}{c}{Second Level} \\

&Acc&Macro-F1&Acc&Macro-F1\\
\hline
\citet{liu-li-2016-recognizing}&57.67&46.13&-&-\\
\citet{chen-etal-2016-implicit}&57.33&45.11&-&-\\

\citet{lan-etal-2017-multi}&57.06&47.29&-&-\\
\citet{ruan-etal-2020-interactively}&58.01&49.45&-&-\\
\citet{xiang-etal-2022-encoding} &60.45&53.14&-&-\\
(BiLSTM)&&&&\\
\citet{xiang-etal-2022-encoding}& 64.04&56.63&-&-\\
(BERT)&&&&\\
Long (RoBERTa-base)&73.32&69.02&63.24&51.80\\
\citet{xiang-etal-2022-connprompt}&74.36&69.91&-&-\\
(RoBERTa-base)&&&&\\
\hline
Ours (RoBERTa-base)&\textbf{75.37}&\textbf{71.19}&\textbf{63.53}&\textbf{52.91}

\\
\hline
\toprule
\end{tabular}
\caption
{Experimental results on PDTB-3.}
\vspace{-0.3cm}
\end{table}

\begin{table}[!tbp]
\centering
\scriptsize
\begin{tabular}{l|cc|cc}
\bottomrule

Model&Comp.&Cont&Exp.&Temp.\\
\hline
\citet{Nguyen2019EmployingTC}&48.44& 56.84 &73.66 &38.60\\
\citet{Guo2020WorkingMN}&43.92& 57.67 &73.45 &36.33\\
\citet{liu2020importancewordsentencerepresentation}&\underline{59.44} &60.98 &77.66 & \underline{50.26}\\
\citet{jiang-etal-2021-just}&55.40& 57.04 &74.76& 41.54\\
\citet{dou-etal-2021-cvae-based}&55.72 &\underline{63.39} &\textbf{80.34} &44.01\\
\citet{long-webber-2022-facilitating}&65.84&63.55&79.17&69.86\\
\citet{Chan2023DiscoPromptPP} (T5-base)&62.55&64.45&78.77&57.41\\
\hline

Ours&\textbf{67.31} &\textbf{66.31} &\underline{78.79} &\textbf{66.21} 
\\
\hline
\toprule
\end{tabular}
\caption
{The results for relation types at level-1 on PDTB-2 in terms of F1 (\%) (top-level multi-class classification).}
\label{table-1}
\end{table}

\begin{table}
\setlength{\belowcaptionskip}{-10pt}
\centering
\scriptsize
\begin{tabular}{l|ccc|c}
\bottomrule

Second-level Label&Liu&Wu&Long&Ours\\
\hline
Temp.Asynchronous&56.18& 56.47 & 59.79&\textbf{60.61}\\
Temp.Synchrony&0.00& 0.00&	\textbf{78.26}&\textbf{78.26} \\
\hline
Cont.Cause&59.60 &64.36 &	65.58&\textbf{66.89} \\
Cont.Pragmatic cause&0.0 &0.0&	0.00 &0.00\\
\hline
Comp.Contrast &59.75& 63.52&	62.63&\textbf{63.76}\\
Comp.Concession& 0.0 &0.0&	0.00 &0\\
\hline
Exp.Conjunction &\textbf{60.17}& 57.91 	&58.35&59.46 \\
Exp.Instantiation& 67.96 &72.60& 	\textbf{73.04}&68.38 \\
Exp.Restatement &53.83 &58.06 &	\textbf{60.00}&53.70 \\
Exp.Alternative  &60.00& 63.46&	53.85&\textbf{66.67} \\
Exp.List &0.0 &8.98 &	\textbf{34.78}&0.00 \\


\hline
\toprule
\end{tabular}
\caption
{The results for relation types at level-2 on PDTB-2 in terms of F1 (\%) (second-level multi-class classification).}
\label{table-4}
\end{table}
\subsection{Parameters Setting and Evaluation Metrics}
\label{sec4.3}
In our monolingual experiments, we use RoBERTa-base \cite{DBLP:journals/corr/abs-1907-11692} as the PLM backbone. We adopt Adam \cite{kingma:adam} with the learning rate of $5e{-}5$ and the batch size of 196 to update the model. The maximum training epoch is set to 10 and the wait patience for early stopping is set to 5 for all models. The temperature $\tau$ is set to 0.1.  For prototype learning, we set the prototype dimension to 128. All experiments are performed with 1× 80GB NVIDIA A100 GPU. Accuracy and Macro-F1 score are used as evaluation metrics. We use the same model and parameters for PDTB-2 and PDTB-3.

\section{Result and Analysis}
\subsection{Main Results and Visualizations}
The results on PDTB-2 and PDTB-3 for Level-1 and Level-2 are given in Table 1 and Table 2 respectively, where the best results are highlighted in bold. Classification performance on PDTB-2 in terms of Macro-F1 for the four general sense types at Level-1 and 11 sense types at Level-2 is shown in Table 3 and Table 4. These results demonstrate better performance than previous systems for both Level-1 and Level-2 classification on both PDTB-2 and PDTB-3, which shows the verbalizer in our work is not inferior to the manual verbalizers proposed by previous work. 




\subsection{Analysis of the Learned Prototypes}
We tried to gain a deeper understanding about the learned prototypes based on our experimental results on PDTB-2 and PDTB-3. 
\begin{figure}[!tbp]
\setlength{\belowcaptionskip}{-8pt}

\centering

\includegraphics[width=6.5cm]{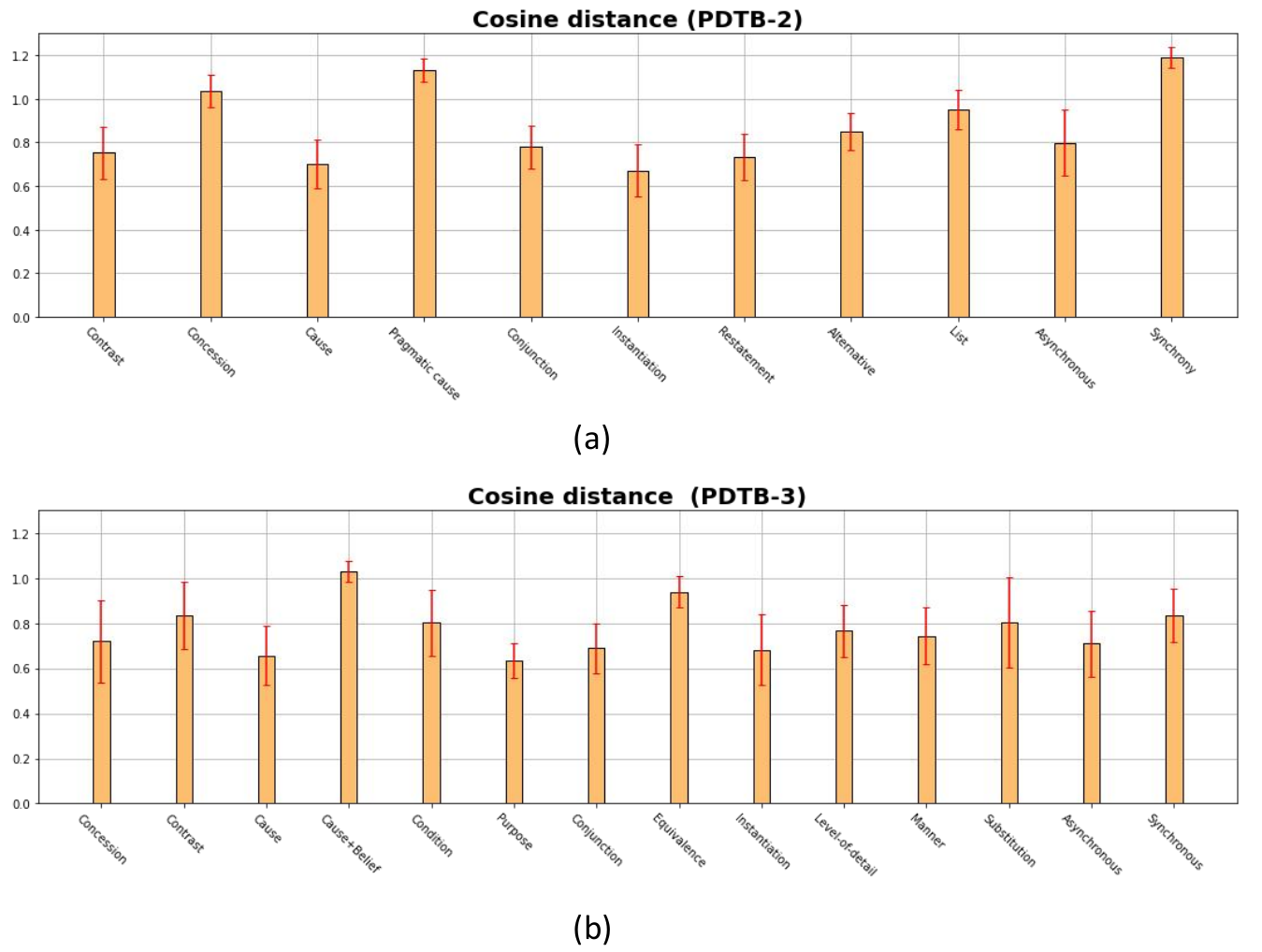}

\caption{Average Cosine Distances between the prototype and all the test examples from the same class with the prototype.}
\label{cos_dis}
\vspace{-0.1cm}
\end{figure}

First, to assess the quality of learned prototypes, we computed the average cosine distances between each prototype and all test examples that are from the same class of the prototype. This provides insights into which prototypes may not have been learned effectively. Larger average cosine distances indicate poorer prototype performance. Figure \ref{cos_dis} illustrates that, when comparing PDTB-2 and PDTB-3, all second-level prototypes in PDTB-3 exhibit superior learning compared to their PDTB-2 counterparts. In the case of PDTB-2, prototypes for ``Synchrony'', ``Pragmatic Cause'', and ``Concession'' demonstrate relatively greater distances from examples annotated with corresponding labels. As for PDTB-3, the prototypes that display room for improvement include ``Cause+belief'', ``Equivalence'', and ``Contrast''. 

Then, we investigated which examples are most similar to the learned prototypes. We got the top ten nearest neighbours for each second-level prototype for PDTB-2 and PDTB-3 in order to assess how many neighbours have the same class as the corresponding prototypes and what labels of the neighbours are if the label is not the class of the prototype. As is shown in Figure \ref{ld_2} and Figure \ref{ld_3}, most of the neighbours have the same class as the prototype, which also verifies the quality of our learned prototypes. Compared with PDTB-2, more neighbours of the prototypes for PDTB-3 belong to the class of the prototypes. 

Upon closer examination of the figure for PDTB-2, an intriguing observation emerges. Specifically, when analyzing the prototype learned for the discourse relation ``Synchrony'', we can see a significant number of its closest neighbors belong to the ``Conjunction'' label. This finding motivated us to further investigate these examples labeled as ``Conjunction''. Then, we found that these examples could potentially have a second label, namely ``Synchrony'', even though it was not annotated by the human annotators. The following is one example:
\begin{figure}[!tbp]
\setlength{\belowcaptionskip}{-8pt}

\centering

\includegraphics[width=7.6cm]{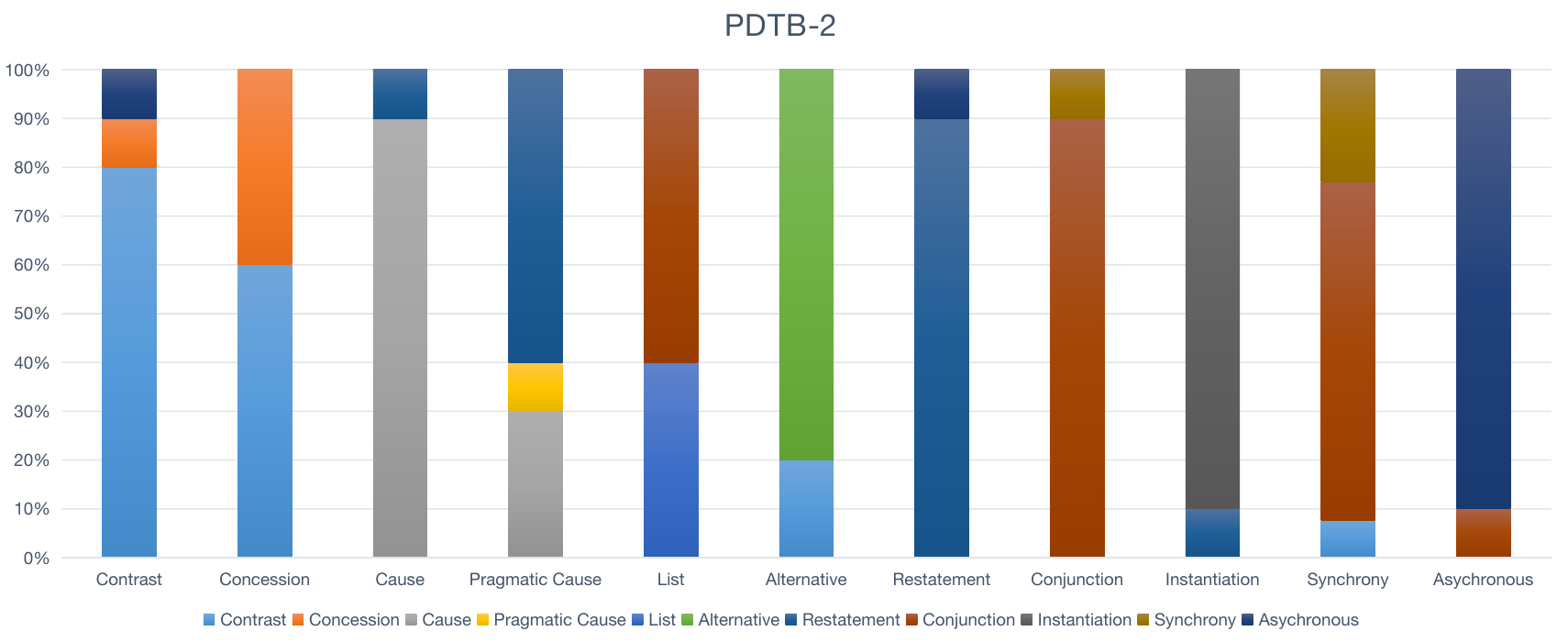}

\caption{Label distribution of the top ten nearest neighbors for each second level prototype in PDTB-2.}
\label{ld_2}
\end{figure}
\begin{figure}[!tbp]
\setlength{\belowcaptionskip}{-8pt}

\centering

\includegraphics[width=7.6cm]{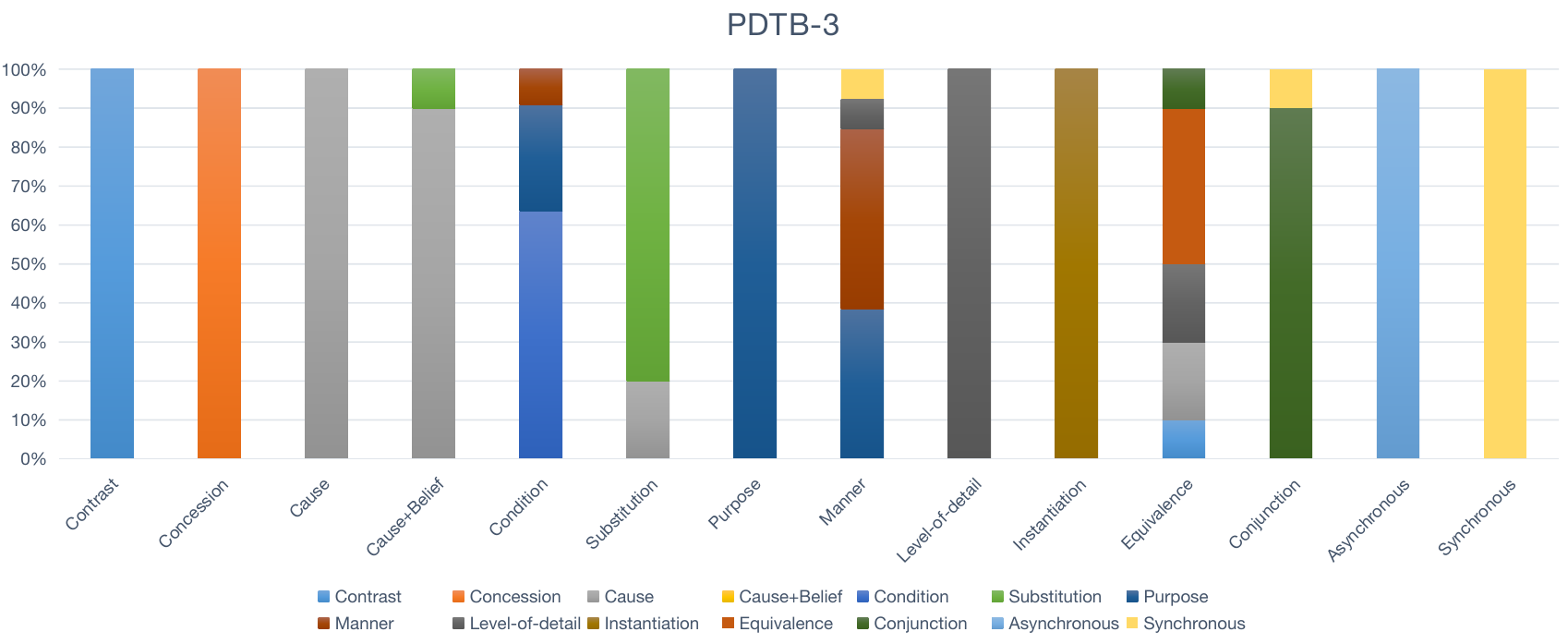}

\caption{Label distribution of the top ten nearest neighbors for each second level prototypes in PDTB-3.}
\label{ld_3}
\vspace{-0.2cm}
\end{figure}


\begin{table*}[!tbp]
\centering
\footnotesize
\begin{tabular}{ll|cc|cc}
\bottomrule
\multirow{2}{*}{Datasets}&\multirow{2}{*}{Model}&\multicolumn{2}{c|}{Top Level}&\multicolumn{2}{c}{Second Level}\\ 


&&Acc&Macro-F1&Acc&Macro-F1\\
\hline

\multirow{4}{*}{PDTB-2}&Ours&\textbf{72.47}&\textbf{69.66}&\textbf{60.73}&\textbf{47.07}\\
&w/o ins\_ins&71.51&68.44&59.19&45.73\\
&w/o pro\_pro&70.83&67.32&58.32&42.10\\
&w/o ins\_ins \& pro\_pro&68.14&63.27&56.02&42.76\\
&w/o label information&72.18&69.07&59.86&45.33\\
\hline 
\multirow{4}{*}{PDTB-3}
&Ours&\textbf{75.37}&\textbf{71.19}&\textbf{63.13}&\textbf{52.91}
\\
&w/o ins\_ins&73.59&69.80&61.31&50.48\\
&w/o pro\_pro&72.77&69.64&61.00&50.27\\
&w/o ins\_ins \& pro\_pro&71.67&67.63&60.08&49.36\\
&w/o label information&74.48&70.52&62.00&51.19\\

\hline
\toprule
\end{tabular}
\setlength{\belowcaptionskip}{-0.2cm}
\caption
{Ablation study on PDTB-2 and PDTB-3. }
\label{abla}
\end{table*}

\begin{enumerate}
    \item [(1)]
[Junk bonds also recovered somewhat, though trading remained stalled.]$_{1}$, [Gold also rose.]$_{2}$. 
\end{enumerate}

The two arguments in the above example are overlapped temporarily. Although the label ``Synchrony'' is not annotated by the annotator, this example could also possess this label. Several factors might result in that. First, annotators may have been instructed to prioritize one label if they encounter instances that could have multiple labels. Second, in the absence of explicit guidance on label selection, annotators might assign only one label to examples that could potentially have two labels. Lastly, the consistency of annotations across the corpus was not thoroughly verified. This phenomenon has implications for the learning of prototypes and their association with each class. We will delve deeper into such cases, analyzing their impacts on systems and how to reduce those impacts in our future work.

\subsection{Ablation Studies}
In order to know the effects of each of the three losses and the effects of the label information provided in our template, we carried ablation studies on both PDTB-2 and PDTB-3.

\begin{table*}[!tbp]
\centering
\footnotesize
\begin{tabular}{l|ccccccc}
\bottomrule
%
Model&German&Lithuanian&Polish&Portuguese&Russian&Turkish\\

\hline
\citet{kurfali-ostling-2019-zero}&39.22&39.32&37.54&39.33&35.50&33.52\\
XLMR-base&41.61&36.00&35.51&34.39&33.11&36.73\\
Ours&\textbf{45.24}&\textbf{41.97}&\textbf{40.49}&\textbf{43.32}&\textbf{37.37}&\textbf{41.12}\\
\hline
\toprule
\end{tabular}
\setlength{\belowcaptionskip}{-0.2cm}
\caption
{Macro-F1 scores (\%) for top level classification on 6 languages when the model is trained on PDTB-3.}
\label{table-6}
\vspace{-0.2cm}

\end{table*}

\begin{figure}[!tbp]
\setlength{\belowcaptionskip}{-8pt}

\centering

\includegraphics[width=7.6cm]{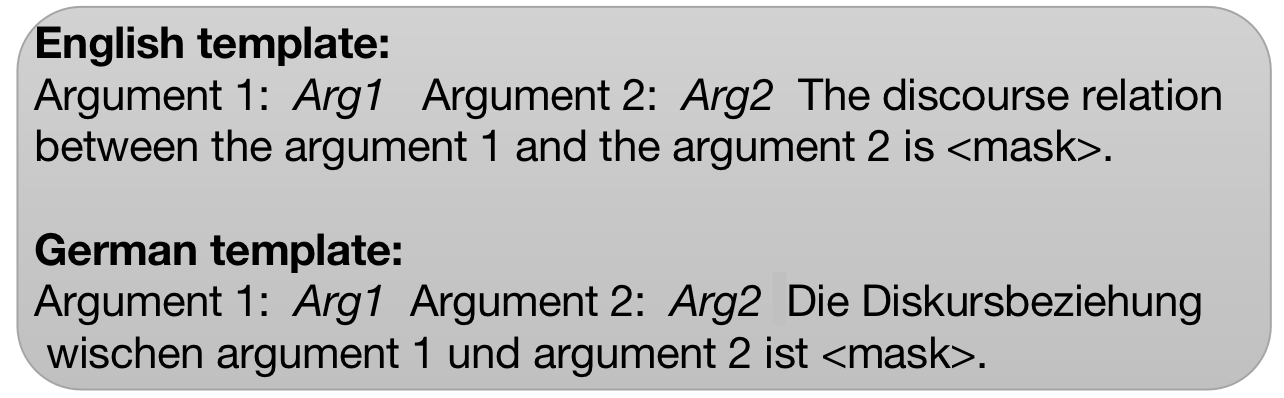}

\caption{One example on constructing our language-specific templates.}
\label{fig7}
\vspace{-0.2cm}
\end{figure}
We firstly did our ablation study by the following: 1) only using instances to prototypes loss; 2) using instance to prototype loss and prototype to prototype loss; 3) using instance to prototype loss and instance to instance loss. From Table \ref{abla}, it can be seen that all the three losses are useful, and the prototype to prototype loss seems to be more important than the instance to instance loss. Overall, combining these three losses can achieve the best results. Besides, as the template we use are injected with the level-1 and level-2 label information, we removed those information in our template to know its effects. We can see from the table that providing these label information is helpful to some degree.




\section{Zero-shot Cross-lingual Transfer Learning}
This section describe how we extend our approach to the zero-shot cross-lingual transfer learning scenario, enabling the models to learn the language-agnostic class features. 
\subsection{Cross-lingual templates}
In order to leverage our approach described in section \ref{sec3} for the target languages and to enhance cross-lingual representation in zero-shot scenarios, a language-specific template is constructed for each target language. Figure \ref{fig7} displays the example about how we construct language-specific template for German.

\begin{figure}[!tbp]
\setlength{\belowcaptionskip}{-8pt}

\centering

\includegraphics[width=7.6cm]{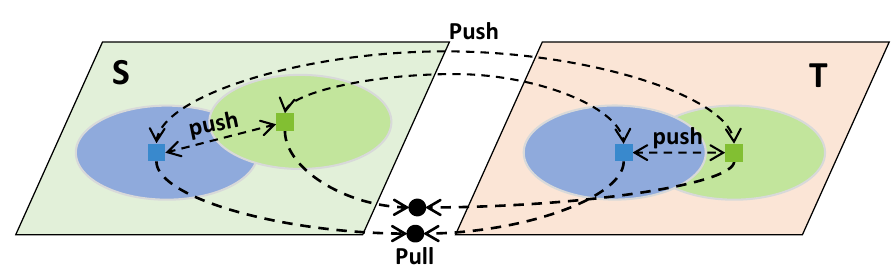}

\caption{Prototype Alignment between the source and the target languages. If the source prototypes and target prototypes are from the same class, we pull them closer, otherwise we push them away.
}
\label{fig8}
\vspace{-0.2cm}
\end{figure}


\subsection{Cross-lingual Prototype Alignment}
The representations of the prototypes for the target languages are trained by using the same methodology we described in section \ref{sec3}. After obtaining all class prototypes for the source and the target language, contrastive learning is employed for adjusting the distance among prototypes in the feature space for class-wise alignment. We bring together the source prototypes and target prototypes if they belong to the same class, while simultaneously pushing away the source and target prototypes if they are from different classes. This behavior is illustrated in Figure \ref{fig8}.


\subsection{Experimental setting}


\subsubsection{Datasets}
We use TED-MDB \cite{zeyrek2019ted} as the target language dataset. TED-MDB is a parallel corpus annotated for discourse relations. It follows the PDTB-3 framework and includes the manual annotations of six TED talks in seven languages(English, Turkish, Portuguese, Polish, German, Russian and Lithuanian). Despite TED-MDB covers many languages, the amount of annotated text per language is very limited. There are around 200 implicit discourse relations for each languages. Due to this, our focus is on the top-level senses. Since TED-MDB follows the PDTB-3 sense hierarchy, PDTB-3 serves as the source language dataset.

\subsubsection{Parameters Setting and Baselines}
For zero-shot cross-lingual experiments, we compare with 2 baselines: (1) vanilla fine tuning the multilingual pre-trained language model on PDTB-3 and tested on target language test set. (2) \citet{kurfali-ostling-2019-zero}, which is the first study about a zero-shot transfer learning for implicit discourse relation recognition. We only used PDTB-3 as the source language dataset, so we compared the results with respect to their models trained on PDTB-3 and tested on TED-MDB. For our method, we use the same parameters presented in the monolingual experiments (Section \ref{sec4.3}). For all cross-lingual experiments (including ours and vanilla fine tuning), we use XLM-RoBERTa base\cite{DBLP:journals/corr/abs-1911-02116}as our multilingual backbone model.




\subsection{Experimental Results}

The results for the top-level on TED-MDB are presented in Table \ref{table-6}. As observed, our method achieves better overall results for all six languages in TED-MDB. Compared with \newcite{kurfali-ostling-2019-zero}, the approach improves on the average Macro-F1 by about 8\% on Turkish. Better performance is achieved in the Portuguese language, with an F1 score of 43.42\%. This result is nearly 10\% higher than the baseline XLMR-based vanilla fine tuning. These outcomes serve as  evidence of the effectiveness of our zero-shot cross-lingual learning methods. 

\section{Conclusions and Future Work}
In this paper, we propose a verbalizer that does not require manual design and leverages the sense hierarchy reflecting the organization of discourse relational senses for the task of implicit discourse relation recognition. Extensive experiments show our method outperforms competitive baselines and effectively supports zero-shot cross-lingual learning for low-resource languages. Additionally, we extend our proposed method to facilitate zero-shot cross-lingual learning for languages with limited annotated data. 

We also observed instances where examples could have multiple valid labels, potentially affecting prototype learning. While not deeply explored here, this will be a focus of future work to understand its impact on system performance to gain a more comprehensive understanding of the complexities involved in this task. Besides, we will consider using other datasets such as TED-CDB\cite{long-etal-2020-ted}, a Chinese discourse bank corpus annotated on TED talks, for cross-lingual transfer learning by leveraging our methods.

\bibliography{acl_latex}

\begin{thebibliography}{46}
\expandafter\ifx\csname natexlab\endcsname\relax\def\natexlab#1{#1}\fi

\bibitem[{Bai and Zhao(2018)}]{bai-zhao-2018-deep}
Hongxiao Bai and Hai Zhao. 2018.
\newblock \href {https://www.aclweb.org/anthology/C18-1048} {Deep enhanced representation for implicit discourse relation recognition}.
\newblock In \emph{Proceedings of the 27th International Conference on Computational Linguistics}, pages 571--583, Santa Fe, New Mexico, USA. Association for Computational Linguistics.

\bibitem[{Chan et~al.(2023)Chan, Liu, Cheng, Li, Song, Wong, and See}]{Chan2023DiscoPromptPP}
Chunkit Chan, Xin Liu, Jiayang Cheng, Zihan Li, Yangqiu Song, Ginny~Y. Wong, and Simon Chong~Wee See. 2023.
\newblock Discoprompt: Path prediction prompt tuning for implicit discourse relation recognition.
\newblock \emph{ArXiv}, abs/2305.03973.

\bibitem[{Chen et~al.(2016{\natexlab{a}})Chen, Zhang, Liu, and Huang}]{Chen2016DiscourseRD}
Jifan Chen, Qi~Zhang, Pengfei Liu, and Xuanjing Huang. 2016{\natexlab{a}}.
\newblock Discourse relations detection via a mixed generative-discriminative framework.
\newblock In \emph{AAAI}.

\bibitem[{Chen et~al.(2016{\natexlab{b}})Chen, Zhang, Liu, Qiu, and Huang}]{chen-etal-2016-implicit}
Jifan Chen, Qi~Zhang, Pengfei Liu, Xipeng Qiu, and Xuanjing Huang. 2016{\natexlab{b}}.
\newblock \href {https://doi.org/10.18653/v1/P16-1163} {Implicit discourse relation detection via a deep architecture with gated relevance network}.
\newblock In \emph{Proceedings of the 54th Annual Meeting of the Association for Computational Linguistics (Volume 1: Long Papers)}, pages 1726--1735, Berlin, Germany. Association for Computational Linguistics.

\bibitem[{Chen et~al.(2020)Chen, Kornblith, Norouzi, and Hinton}]{pmlr-v119-chen20j}
Ting Chen, Simon Kornblith, Mohammad Norouzi, and Geoffrey Hinton. 2020.
\newblock \href {https://proceedings.mlr.press/v119/chen20j.html} {A simple framework for contrastive learning of visual representations}.
\newblock In \emph{Proceedings of the 37th International Conference on Machine Learning}, volume 119 of \emph{Proceedings of Machine Learning Research}, pages 1597--1607. PMLR.

\bibitem[{Conneau et~al.(2019)Conneau, Khandelwal, Goyal, Chaudhary, Wenzek, Guzm{\'{a}}n, Grave, Ott, Zettlemoyer, and Stoyanov}]{DBLP:journals/corr/abs-1911-02116}
Alexis Conneau, Kartikay Khandelwal, Naman Goyal, Vishrav Chaudhary, Guillaume Wenzek, Francisco Guzm{\'{a}}n, Edouard Grave, Myle Ott, Luke Zettlemoyer, and Veselin Stoyanov. 2019.
\newblock \href {http://arxiv.org/abs/1911.02116} {Unsupervised cross-lingual representation learning at scale}.
\newblock \emph{CoRR}, abs/1911.02116.

\bibitem[{Cui et~al.(2022)Cui, Hu, Ding, Huang, and Liu}]{Cui2022PrototypicalVF}
Ganqu Cui, Shengding Hu, Ning Ding, Longtao Huang, and Zhiyuan Liu. 2022.
\newblock Prototypical verbalizer for prompt-based few-shot tuning.
\newblock In \emph{Annual Meeting of the Association for Computational Linguistics}.

\bibitem[{Dai and Huang(2019)}]{dai-huang-2019-regularization}
Zeyu Dai and Ruihong Huang. 2019.
\newblock \href {https://doi.org/10.18653/v1/D19-1295} {A regularization approach for incorporating event knowledge and coreference relations into neural discourse parsing}.
\newblock In \emph{Proceedings of the 2019 Conference on Empirical Methods in Natural Language Processing and the 9th International Joint Conference on Natural Language Processing (EMNLP-IJCNLP)}, pages 2976--2987, Hong Kong, China. Association for Computational Linguistics.

\bibitem[{Ding et~al.(2021)Ding, Wang, Fu, Xu, Wang, Xie, Shen, Huang, Zheng, and Zhang}]{ding2021prototypical}
Ning Ding, Xiaobin Wang, Yao Fu, Guangwei Xu, Rui Wang, Pengjun Xie, Ying Shen, Fei Huang, Hai-Tao Zheng, and Rui Zhang. 2021.
\newblock \href {http://arxiv.org/abs/2103.11647} {Prototypical representation learning for relation extraction}.

\bibitem[{Dou et~al.(2021)Dou, Hong, Sun, and Zhou}]{dou-etal-2021-cvae-based}
Zujun Dou, Yu~Hong, Yu~Sun, and Guodong Zhou. 2021.
\newblock \href {https://doi.org/10.18653/v1/2021.findings-emnlp.110} {{CVAE}-based re-anchoring for implicit discourse relation classification}.
\newblock In \emph{Findings of the Association for Computational Linguistics: EMNLP 2021}, pages 1275--1283, Punta Cana, Dominican Republic. Association for Computational Linguistics.

\bibitem[{Gao et~al.(2021)Gao, Fisch, and Chen}]{gao-etal-2021-making}
Tianyu Gao, Adam Fisch, and Danqi Chen. 2021.
\newblock \href {https://doi.org/10.18653/v1/2021.acl-long.295} {Making pre-trained language models better few-shot learners}.
\newblock In \emph{Proceedings of the 59th Annual Meeting of the Association for Computational Linguistics and the 11th International Joint Conference on Natural Language Processing (Volume 1: Long Papers)}, pages 3816--3830, Online. Association for Computational Linguistics.

\bibitem[{Guo et~al.(2020)Guo, He, Dang, and Wang}]{Guo2020WorkingMN}
Fengyu Guo, Ruifang He, Jianwu Dang, and Jian Wang. 2020.
\newblock Working memory-driven neural networks with a novel knowledge enhancement paradigm for implicit discourse relation recognition.
\newblock In \emph{AAAI}.

\bibitem[{Huang and Kurohashi(2021)}]{huang-kurohashi-2021-extractive}
Yin~Jou Huang and Sadao Kurohashi. 2021.
\newblock \href {https://doi.org/10.18653/v1/2021.eacl-main.265} {Extractive summarization considering discourse and coreference relations based on heterogeneous graph}.
\newblock In \emph{Proceedings of the 16th Conference of the European Chapter of the Association for Computational Linguistics: Main Volume}, pages 3046--3052, Online. Association for Computational Linguistics.

\bibitem[{Jiang et~al.(2021)Jiang, Fan, Chu, Li, and Zhu}]{jiang-etal-2021-just}
Feng Jiang, Yaxin Fan, Xiaomin Chu, Peifeng Li, and Qiaoming Zhu. 2021.
\newblock \href {https://doi.org/10.18653/v1/2021.emnlp-main.187} {Not just classification: Recognizing implicit discourse relation on joint modeling of classification and generation}.
\newblock In \emph{Proceedings of the 2021 Conference on Empirical Methods in Natural Language Processing}, pages 2418--2431, Online and Punta Cana, Dominican Republic. Association for Computational Linguistics.

\bibitem[{Khosla et~al.(2020)Khosla, Teterwak, Wang, Sarna, Tian, Isola, Maschinot, Liu, and Krishnan}]{khosla2020supervised}
Prannay Khosla, Piotr Teterwak, Chen Wang, Aaron Sarna, Yonglong Tian, Phillip Isola, Aaron Maschinot, Ce~Liu, and Dilip Krishnan. 2020.
\newblock Supervised contrastive learning.
\newblock \emph{arXiv preprint arXiv:2004.11362}.

\bibitem[{Kingma and Ba(2015)}]{kingma:adam}
Diederick~P Kingma and Jimmy Ba. 2015.
\newblock Adam: A method for stochastic optimization.
\newblock In \emph{International Conference on Learning Representations (ICLR)}.

\bibitem[{Kishimoto et~al.(2020)Kishimoto, Murawaki, and Kurohashi}]{Kishimoto2020AdaptingBT}
Yudai Kishimoto, Yugo Murawaki, and Sadao Kurohashi. 2020.
\newblock Adapting bert to implicit discourse relation classification with a focus on discourse connectives.
\newblock In \emph{LREC}.

\bibitem[{Kurfal{\i} and {\"O}stling(2019)}]{kurfali-ostling-2019-zero}
Murathan Kurfal{\i} and Robert {\"O}stling. 2019.
\newblock \href {https://doi.org/10.18653/v1/W19-5927} {Zero-shot transfer for implicit discourse relation classification}.
\newblock In \emph{Proceedings of the 20th Annual SIGdial Meeting on Discourse and Dialogue}, pages 226--231, Stockholm, Sweden. Association for Computational Linguistics.

\bibitem[{Lan et~al.(2017)Lan, Wang, Wu, Niu, and Wang}]{lan-etal-2017-multi}
Man Lan, Jianxiang Wang, Yuanbin Wu, Zheng-Yu Niu, and Haifeng Wang. 2017.
\newblock \href {https://doi.org/10.18653/v1/D17-1134} {Multi-task attention-based neural networks for implicit discourse relationship representation and identification}.
\newblock In \emph{Proceedings of the 2017 Conference on Empirical Methods in Natural Language Processing}, pages 1299--1308, Copenhagen, Denmark. Association for Computational Linguistics.

\bibitem[{Liu et~al.(2020)Liu, Ou, Song, and Jiang}]{liu2020importancewordsentencerepresentation}
Xin Liu, Jiefu Ou, Yangqiu Song, and Xin Jiang. 2020.
\newblock \href {http://arxiv.org/abs/2004.12617} {On the importance of word and sentence representation learning in implicit discourse relation classification}.

\bibitem[{Liu and Li(2016)}]{liu-li-2016-recognizing}
Yang Liu and Sujian Li. 2016.
\newblock \href {https://doi.org/10.18653/v1/D16-1130} {Recognizing implicit discourse relations via repeated reading: Neural networks with multi-level attention}.
\newblock In \emph{Proceedings of the 2016 Conference on Empirical Methods in Natural Language Processing}, pages 1224--1233, Austin, Texas. Association for Computational Linguistics.

\bibitem[{Liu et~al.(2019)Liu, Ott, Goyal, Du, Joshi, Chen, Levy, Lewis, Zettlemoyer, and Stoyanov}]{DBLP:journals/corr/abs-1907-11692}
Yinhan Liu, Myle Ott, Naman Goyal, Jingfei Du, Mandar Joshi, Danqi Chen, Omer Levy, Mike Lewis, Luke Zettlemoyer, and Veselin Stoyanov. 2019.
\newblock \href {http://arxiv.org/abs/1907.11692} {Roberta: {A} robustly optimized {BERT} pretraining approach}.
\newblock \emph{CoRR}, abs/1907.11692.

\bibitem[{Long and Webber(2022)}]{long-webber-2022-facilitating}
Wanqiu Long and Bonnie Webber. 2022.
\newblock \href {https://aclanthology.org/2022.emnlp-main.734} {Facilitating contrastive learning of discourse relational senses by exploiting the hierarchy of sense relations}.
\newblock In \emph{Proceedings of the 2022 Conference on Empirical Methods in Natural Language Processing}, pages 10704--10716, Abu Dhabi, United Arab Emirates. Association for Computational Linguistics.

\bibitem[{Long et~al.(2020)Long, Webber, and Xiong}]{long-etal-2020-ted}
Wanqiu Long, Bonnie Webber, and Deyi Xiong. 2020.
\newblock \href {https://doi.org/10.18653/v1/2020.emnlp-main.223} {{TED}-{CDB}: A large-scale {C}hinese discourse relation dataset on {TED} talks}.
\newblock In \emph{Proceedings of the 2020 Conference on Empirical Methods in Natural Language Processing (EMNLP)}, pages 2793--2803, Online. Association for Computational Linguistics.

\bibitem[{Nguyen et~al.(2019)Nguyen, Linh, Than, and Nguyen}]{Nguyen2019EmployingTC}
Linh~The Nguyen, Ngo~Van Linh, Khoat Than, and Thien~Huu Nguyen. 2019.
\newblock Employing the correspondence of relations and connectives to identify implicit discourse relations via label embeddings.
\newblock In \emph{ACL}.

\bibitem[{Prasad et~al.(2008)Prasad, Dinesh, Lee, Miltsakaki, Robaldo, Joshi, and Webber}]{prasad-etal-2008-penn}
Rashmi Prasad, Nikhil Dinesh, Alan Lee, Eleni Miltsakaki, Livio Robaldo, Aravind Joshi, and Bonnie Webber. 2008.
\newblock \href {http://www.lrec-conf.org/proceedings/lrec2008/pdf/754_paper.pdf} {The {P}enn discourse {T}ree{B}ank 2.0.}
\newblock In \emph{Proceedings of the Sixth International Conference on Language Resources and Evaluation ({LREC}'08)}, Marrakech, Morocco. European Language Resources Association (ELRA).

\bibitem[{Pyatkin et~al.(2020)Pyatkin, Klein, Tsarfaty, and Dagan}]{pyatkin-etal-2020-qadiscourse}
Valentina Pyatkin, Ayal Klein, Reut Tsarfaty, and Ido Dagan. 2020.
\newblock \href {https://doi.org/10.18653/v1/2020.emnlp-main.224} {{QAD}iscourse - {D}iscourse {R}elations as {QA} {P}airs: {R}epresentation, {C}rowdsourcing and {B}aselines}.
\newblock In \emph{Proceedings of the 2020 Conference on Empirical Methods in Natural Language Processing (EMNLP)}, pages 2804--2819, Online. Association for Computational Linguistics.

\bibitem[{Qin et~al.(2016)Qin, Zhang, and Zhao}]{shallow}
Lianhui Qin, Zhisong Zhang, and Hai Zhao. 2016.
\newblock \href {https://doi.org/10.18653/v1/K16-2010} {Shallow discourse parsing using convolutional neural network}.
\newblock In \emph{Proceedings of the {C}o{NLL}-16 shared task}, pages 70--77, Berlin, Germany. Association for Computational Linguistics.

\bibitem[{Ruan et~al.(2020)Ruan, Hong, Xu, Huang, Zhou, and Zhang}]{ruan-etal-2020-interactively}
Huibin Ruan, Yu~Hong, Yang Xu, Zhen Huang, Guodong Zhou, and Min Zhang. 2020.
\newblock \href {https://doi.org/10.18653/v1/2020.coling-main.282} {Interactively-propagative attention learning for implicit discourse relation recognition}.
\newblock In \emph{Proceedings of the 28th International Conference on Computational Linguistics}, pages 3168--3178, Barcelona, Spain (Online). International Committee on Computational Linguistics.

\bibitem[{Schick and Sch{\"u}tze(2021)}]{schick-schutze-2021-exploiting}
Timo Schick and Hinrich Sch{\"u}tze. 2021.
\newblock \href {https://doi.org/10.18653/v1/2021.eacl-main.20} {Exploiting cloze-questions for few-shot text classification and natural language inference}.
\newblock In \emph{Proceedings of the 16th Conference of the European Chapter of the Association for Computational Linguistics: Main Volume}, pages 255--269, Online. Association for Computational Linguistics.

\bibitem[{Shi and Demberg(2019{\natexlab{a}})}]{shi-demberg-2019-learning}
Wei Shi and Vera Demberg. 2019{\natexlab{a}}.
\newblock \href {https://doi.org/10.18653/v1/W19-0416} {Learning to explicitate connectives with {S}eq2{S}eq network for implicit discourse relation classification}.
\newblock In \emph{Proceedings of the 13th International Conference on Computational Semantics - Long Papers}, pages 188--199, Gothenburg, Sweden. Association for Computational Linguistics.

\bibitem[{Shi and Demberg(2019{\natexlab{b}})}]{shi-demberg-2019-next}
Wei Shi and Vera Demberg. 2019{\natexlab{b}}.
\newblock \href {https://doi.org/10.18653/v1/D19-1586} {Next sentence prediction helps implicit discourse relation classification within and across domains}.
\newblock In \emph{Proceedings of the 2019 Conference on Empirical Methods in Natural Language Processing and the 9th International Joint Conference on Natural Language Processing (EMNLP-IJCNLP)}, pages 5790--5796, Hong Kong, China. Association for Computational Linguistics.

\bibitem[{Shin et~al.(2020)Shin, Razeghi, Logan~IV, Wallace, and Singh}]{shin-etal-2020-autoprompt}
Taylor Shin, Yasaman Razeghi, Robert~L. Logan~IV, Eric Wallace, and Sameer Singh. 2020.
\newblock \href {https://doi.org/10.18653/v1/2020.emnlp-main.346} {{A}uto{P}rompt: {E}liciting {K}nowledge from {L}anguage {M}odels with {A}utomatically {G}enerated {P}rompts}.
\newblock In \emph{Proceedings of the 2020 Conference on Empirical Methods in Natural Language Processing (EMNLP)}, pages 4222--4235, Online. Association for Computational Linguistics.

\bibitem[{Snell et~al.(2017)Snell, Swersky, and Zemel}]{snell2017prototypical}
Jake Snell, Kevin Swersky, and Richard~S. Zemel. 2017.
\newblock \href {http://arxiv.org/abs/1703.05175} {Prototypical networks for few-shot learning}.

\bibitem[{Song et~al.(2022)Song, Huang, Xue, and Hu}]{song2022supervised}
Xiaohui Song, Longtao Huang, Hui Xue, and Songlin Hu. 2022.
\newblock \href {http://arxiv.org/abs/2210.08713} {Supervised prototypical contrastive learning for emotion recognition in conversation}.

\bibitem[{Tang et~al.(2021)Tang, Lin, Liao, Lu, Han, Sun, Xie, and Xu}]{tang-etal-2021-discourse}
Jialong Tang, Hongyu Lin, Meng Liao, Yaojie Lu, Xianpei Han, Le~Sun, Weijian Xie, and Jin Xu. 2021.
\newblock \href {https://doi.org/10.18653/v1/2021.acl-long.60} {From discourse to narrative: Knowledge projection for event relation extraction}.
\newblock In \emph{Proceedings of the 59th Annual Meeting of the Association for Computational Linguistics and the 11th International Joint Conference on Natural Language Processing (Volume 1: Long Papers)}, pages 732--742, Online. Association for Computational Linguistics.

\bibitem[{Webber et~al.(2019)Webber, Prasad, Lee, and Joshi}]{webber2019penn}
Bonnie Webber, Rashmi Prasad, Alan Lee, and Aravind Joshi. 2019.
\newblock The penn discourse treebank 3.0 annotation manual.

\bibitem[{Wu et~al.(2021{\natexlab{a}})Wu, Cao, Ge, Liu, Zhang, and Su}]{wu2021label}
Changxing Wu, Liuwen Cao, Yubin Ge, Yang Liu, Min Zhang, and Jinsong Su. 2021{\natexlab{a}}.
\newblock A label dependence-aware sequence generation model for multi-level implicit discourse relation recognition.
\newblock \emph{arXiv preprint arXiv:2112.11740}.

\bibitem[{Wu et~al.(2021{\natexlab{b}})Wu, Cao, Ge, Liu, Zhang, and Su}]{Wu2021ALD}
Changxing Wu, Liuwen Cao, Yubin Ge, Yang Liu, Min Zhang, and Jinsong Su. 2021{\natexlab{b}}.
\newblock A label dependence-aware sequence generation model for multi-level implicit discourse relation recognition.
\newblock \emph{ArXiv}, abs/2112.11740.

\bibitem[{Wu et~al.(2022)Wu, Cao, Ge, Liu, Zhang, and Su}]{Wu2022ALD}
Changxing Wu, Liuwen Cao, Yubin Ge, Yang Liu, Min Zhang, and Jinsong Su. 2022.
\newblock A label dependence-aware sequence generation model for multi-level implicit discourse relation recognition.
\newblock In \emph{AAAI}.

\bibitem[{Xiang et~al.(2022{\natexlab{a}})Xiang, Wang, Dai, and Mo}]{xiang-etal-2022-encoding}
Wei Xiang, Bang Wang, Lu~Dai, and Yijun Mo. 2022{\natexlab{a}}.
\newblock \href {https://doi.org/10.18653/v1/2022.findings-acl.256} {Encoding and fusing semantic connection and linguistic evidence for implicit discourse relation recognition}.
\newblock In \emph{Findings of the Association for Computational Linguistics: ACL 2022}, pages 3247--3257, Dublin, Ireland. Association for Computational Linguistics.

\bibitem[{Xiang et~al.(2022{\natexlab{b}})Xiang, Wang, Dai, and Wang}]{xiang-etal-2022-connprompt}
Wei Xiang, Zhenglin Wang, Lu~Dai, and Bang Wang. 2022{\natexlab{b}}.
\newblock \href {https://aclanthology.org/2022.coling-1.75} {{C}onn{P}rompt: Connective-cloze prompt learning for implicit discourse relation recognition}.
\newblock In \emph{Proceedings of the 29th International Conference on Computational Linguistics}, pages 902--911, Gyeongju, Republic of Korea. International Committee on Computational Linguistics.

\bibitem[{Zeyrek et~al.(2019)Zeyrek, Mendes, Grishina, Kurfali, Gibbon, and Ogrodniczuk}]{zeyrek2019ted}
Deniz Zeyrek, Amalia Mendes, Yulia Grishina, Murathan Kurfali, Samuel Gibbon, and Maciej Ogrodniczuk. 2019.
\newblock Ted multilingual discourse bank (ted-mdb): a parallel corpus annotated in the pdtb style.
\newblock \emph{Language Resources and Evaluation}, pages 1--38.

\bibitem[{Zhao et~al.(2023)Zhao, Ouyang, Yu, Wu, and Li}]{zhao-etal-2023-pre}
Xuandong Zhao, Siqi Ouyang, Zhiguo Yu, Ming Wu, and Lei Li. 2023.
\newblock \href {https://doi.org/10.18653/v1/2023.acl-long.869} {Pre-trained language models can be fully zero-shot learners}.
\newblock In \emph{Proceedings of the 61st Annual Meeting of the Association for Computational Linguistics (Volume 1: Long Papers)}, pages 15590--15606, Toronto, Canada. Association for Computational Linguistics.

\bibitem[{Zhou et~al.(2022)Zhou, Lan, Wu, Chen, and Ma}]{zhou-etal-2022-prompt-based}
Hao Zhou, Man Lan, Yuanbin Wu, Yuefeng Chen, and Meirong Ma. 2022.
\newblock \href {https://aclanthology.org/2022.findings-emnlp.282} {Prompt-based connective prediction method for fine-grained implicit discourse relation recognition}.
\newblock In \emph{Findings of the Association for Computational Linguistics: EMNLP 2022}, pages 3848--3858, Abu Dhabi, United Arab Emirates. Association for Computational Linguistics.

\bibitem[{Zhou et~al.(2023)Zhou, Li, Bing, Cambria, and Miao}]{zhou2023improving}
Ran Zhou, Xin Li, Lidong Bing, Erik Cambria, and Chunyan Miao. 2023.
\newblock \href {http://arxiv.org/abs/2305.13628} {Improving self-training for cross-lingual named entity recognition with contrastive and prototype learning}.

\end{thebibliography}

\end{document}